\documentclass[twoside]{article}

\usepackage[accepted]{aistats2019}
\usepackage{graphicx}
\usepackage{times}  
\usepackage{helvet}  
\usepackage{courier}  
\usepackage{url}  
\usepackage{amsmath}
\usepackage{amssymb}
\usepackage{algorithmic}
\usepackage{algorithm}
\usepackage{color}
\begin{document}

%

%
\newtheorem{proposition}{Proposition}
\renewcommand{\algorithmicrequire}{\textbf{Input:}}
\renewcommand{\algorithmicensure}{\textbf{Output:}}

\twocolumn[

\aistatstitle{Binary Space Partitioning Forests}

\aistatsauthor{ Xuhui Fan \And Bin Li \And  Scott A. Sisson }

\aistatsaddress{School of Mathematics and Statistics\\University of New South Wales\\xuhui.fan@unsw.edu.au \And  School of Computer Science\\Fudan University\\libin@fudan.edu.cn \And School of Mathematics and Statistics\\University of New South Wales\\scott.sisson@unsw.edu.au} ]

\begin{abstract}
The Binary Space Partitioning~(BSP)-Tree process is proposed to produce flexible 2-D partition structures which are originally used as a Bayesian nonparametric prior for relational modelling. It can hardly be applied to other learning tasks such as regression trees because extending the BSP-Tree process to a higher dimensional space is nontrivial. This paper is the first attempt to extend the BSP-Tree process to a $d$-dimensional ($d>2$) space. We propose to generate a cutting hyperplane, which is assumed to be parallel to $d-2$ dimensions, to cut each node in the $d$-dimensional BSP-tree. By designing a subtle strategy to sample two free dimensions from $d$ dimensions, the extended BSP-Tree process can inherit the essential self-consistency property from the original version. Based on the extended BSP-Tree process, an ensemble model, which is named the BSP-Forest, is further developed for regression tasks. Thanks to the retained self-consistency property, we can thus significantly reduce the geometric calculations in the inference stage. Compared to its counterpart, the Mondrian Forest, the BSP-Forest can achieve similar performance with fewer cuts due to its flexibility. The BSP-Forest also outperforms other (Bayesian) regression forests on a number of real-world data sets.
\end{abstract}

\section{Introduction}
Several machine learning methods, such as decision trees for regression and relational modelling for identifying interaction patterns, are concerned with space partitioning strategies to identify meaningful ``blocks'' in a product space. Models may be fitted to the data in each block, within which the data will exhibit certain types of homogeneity. These techniques have found application in relational modeling~\cite{kemp2006learning,airoldi2009mixed}, community detection~\cite{nowicki2001estimation,karrer2011stochastic}, collaborative filtering~\cite{porteous2008multi, Li_transfer_2009}, and random forests~\cite{LakRoyTeh2014a} -- This space-partitioning strategy has shown its promising prospect in real-world applications. As a result, a number of structured space-partitioning priors have been developed, including the Mondrian process~\cite{roy2009mondrian, roy2007learning, roy2011thesis}, the Rectangular Tiling process~\cite{nakano2014rectangular}, the Ostomachion process~\cite{xuhui2016OstomachionProcess}, and the Binary Space Partitioning~(BSP)-Tree process~\cite{pmlr-v84-fan18b}. Other strategies are also developed to complete the task, e.g., the Rectangular Bounding Process~\cite{NIPS2018_RBP} is recently proposed to use a bounding strategy to partition the space and it claims to obtain a parsimonious result. 

Among the aforementioned approaches, the BSP-Tree process~\cite{pmlr-v84-fan18b} is an efficient way to partition the two-dimensional space. Instead of axis-aligned cuts adopted in most conventional approaches~\cite{kemp2006learning,roy2009mondrian,nakano2014rectangular}, the BSP-Tree process uses angled cuts to better describe the potential dependency between each dimension. The BSP-tree process is attractive because it is self-consistent, which ensures distributional invariance while restricting the process from a larger domain to a smaller one. However, as the BSP-Tree process is originally proposed for relational modelling, which only requires partitions on a two-dimensional space, it can hardly be extended to a $d$-dimensional ($d>2$) space. This restriction prohibits its applications to decision tree-style
models, which usually consist of more than two dimensions of features. Unfortunately, we cannot straightforwardly extend the BSP-Tree process to a higher dimensional space because it would violate the self-consistency. It is possible to consider a proper extension that is able to retain the self-consistency property, however, for a $d$-dimensional space, the hyperplane usually lies in the $(d-1)$-dimensional space and the measure of all potential cutting hyperplanes involves complicated integrals which would incur unacceptable calculations for a machine learning task. 

In this work, we make the first endeavor to extend the domain of the BSP-Tree process to $d$-dimensional ($d>2$) space while still keeping its self-consistency. To simplify the process, we propose to generate a cutting hyperplane, which is assumed to be parallel to $d-2$ dimensions, to cut each node in the $d$-dimensional BSP-tree. That is to say, each cutting hyperplane is allowed to have two degrees (dimensions) of freedom during the generative process. The current node, which is a convex polyhedron, is first projected onto a pair of dimensions -- The pair is sampled in proportion to the perimeter of its projection onto the sampled pair of dimensions among all $1/2\cdot d(d-1)$ possible configurations. The subsequent cut on this node is then generated on the projected convex polygon through the same way of the BSP-Tree process and the rest dimensions of the cutting hyperplane are parallel to the other dimensions. This geometrically simple construction permits a flexible multi-dimensional extension to the BSP-tree process that provably retains the self-consistency property.

Our second contribution is to construct an ensemble of BSP-trees -- the {\em BSP-forest} -- to enable complex and flexible regression modelling.
In contrast to Bayesian additive regression trees (BART)~\cite{chipman2010bart} and the Mondrian Forest ~\cite{LakRoyTeh2014a}, which only implement node cuts in  a single dimension to generate the tree structure, the BSP-forest uses two dimensions jointly to form a hyperplane cut in the feature space. As a result, the BSP-forest can achieve higher performance with a lower-depth hierarchical tree structure. In addition, because it consists of multiple BST-trees who select different pairs of dimensions to describe the data in a local region, the BSP-Forest is able to capture all-round pairwise dimensional dependence. 

In the inference stage, an efficient Particle Gibbs sampler is developed to infer the BSP-Forest. Particularly, instead of cutting the entire node, we can further simplify the partitioning process by only conducting the hyperplane cut in the convex hull of the training data within the node  to circumvent the complicated polyhedron related evaluation. Thanks to the self-consistency of the proposed extended BSP-Tree process, these two strategies lead to the same equilibrium distribution; we can thus largely simplify the sampling procedure and improve the efficiency.
 
The effectiveness of the proposed BSP-Forest is validated through an extensive study on the famous Friedman's function and then five real-world data sets. Compared to its counterpart, the Mondrian Forest, the BSP-Forest can achieve similar performance with fewer cuts due to its flexibility. The BSP-Forest also outperforms other (Bayesian) regression forests on a number of real-world data sets.



\begin{figure}[t]
\centering
\includegraphics[width =  0.4 \textwidth]{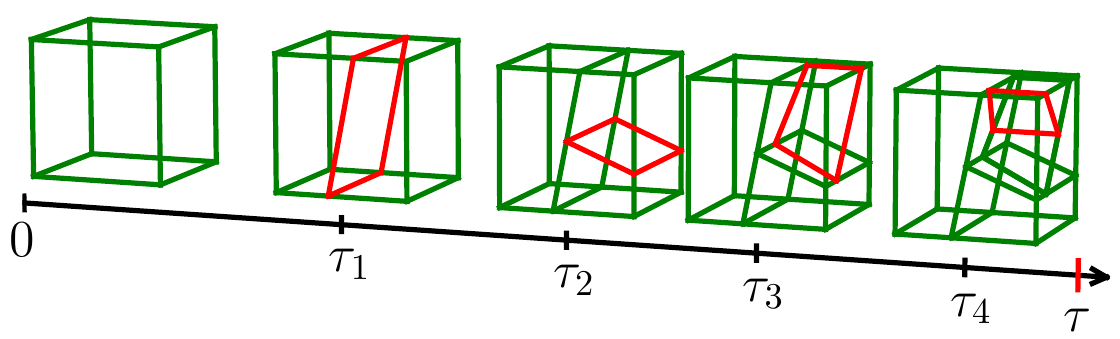}
\caption{A realization of one branch of the extended ($3$-dimensional) BSP-Tree process with budget $\tau$. Each red-line constituted polygon denotes the new cutting hyperplane generated at that time.}
\label{fig:generativeprocesspartition}
\end{figure}

\section{Preliminary: The BSP-Tree Process}

The BSP-tree process~\cite{pmlr-v84-fan18b} generates partitions, $\boxplus$, on an arbitrary two-dimensional complex polygon $\square$.
%
The BSP-tree process is an almost surely right-continuous Markov jump process on $(0, \tau]$, where $\tau>0$ is a pre-fixed budget. Let $\boxplus_t$ denote the BSP-tree partition at time $t$. $\boxplus_t$ is a hierarchical partition of $\square$, which can be represented as a triplet $(\boxplus, \pmb{\theta}, \pmb{u})$. $\boxplus$ is a finite binary space partitioning tree, and $\pmb{\theta}=(\theta_j)$ and $\pmb{u} = (\pmb{u}_j)$ respectively specify, for each intermediate node $j$ of the tree, the orthogonal slope ($\theta_j$) for the cutting line and the cut position ($\pmb{u}_j$) on the projected segment of the node. 

The probability density function of $\theta_j$ is proportional to the length of the projected segment $\pmb{l}(\theta_j)$, and the cut position $\pmb{u}_j$ is uniformly distributed on $\pmb{l}(\theta_j)$. 
$\boxplus_t$ represents a hierarchical partition of $\square$ into the root node ($\square_{\epsilon}=\square$; the base of the domain), intermediate nodes ($\square_{j}$) and terminal nodes.
Intermediate nodes $\square_{j}$ are generated from their parental node through a node split and are themselves split into two child nodes through the cutting line
\begin{align}
    &\{\pmb{x}\in\square_j:(\pmb{x}-\pmb{u}_j)(1;\tan\theta_j)^{\top}<0\},\nonumber\\ &\{\pmb{x}\in\square_j:(\pmb{x}-\pmb{u}_j)(1;\tan\theta_j)^{\top}>0\}.\nonumber
\end{align}
Terminal nodes are the final generated blocks, which do not contain cutting lines. 


Defining the time for the $l$-th cut as $t=\tau_l$,
the incremental cut time $\tau_{l}-\tau_{l-1}$ depends only on the partitions at time $\tau_{l-1}$. Given an existing partition $\boxplus_{\tau_{l-1}}$, the time to the next cut
follows an Exponential distribution:
\begin{align} \label{elaping_time}
    (\tau_{l}-\tau_{l-1})|\boxplus_{\tau_{l-1}}\sim \textrm{Exp}(\sum_{k=1}^lPE(\square_{\tau_{l-1}}^{(k)}))
\end{align}
where $PE(\square_{\tau_{l-1}}^{(k)})$ denotes the perimeter of the $k$-th block in partition $\boxplus_{\tau_{l-1}}$. Each cut divides one block (chosen with probability proportional to its perimeter) into two new blocks and forms a new partition. If the time index $\tau_l$  of the new cut exceeds the budget $\tau$, the BSP-tree process terminates and returns the partition $\boxplus_{\tau_{l-1}}$ as the final realisation. 

\paragraph{Self-consistency} Self-consistency of the BSP-Tree process refers to that, when restricting the BSP-tree process on a convex polygon $\square$ to any sub-region $\triangle\subseteq\square$, the resulting partitioning on the sub-region is distributed as if it is directly generated on $\triangle$ through the BSP-tree's generative process. This property can be usefully exploited in settings such as online learning and domain expansion. 
See \cite{pmlr-v84-fan18b} for more detailed description of the BSP-tree process.

\begin{figure*}[t]
\centering
\includegraphics[width =  0.75 \textwidth]{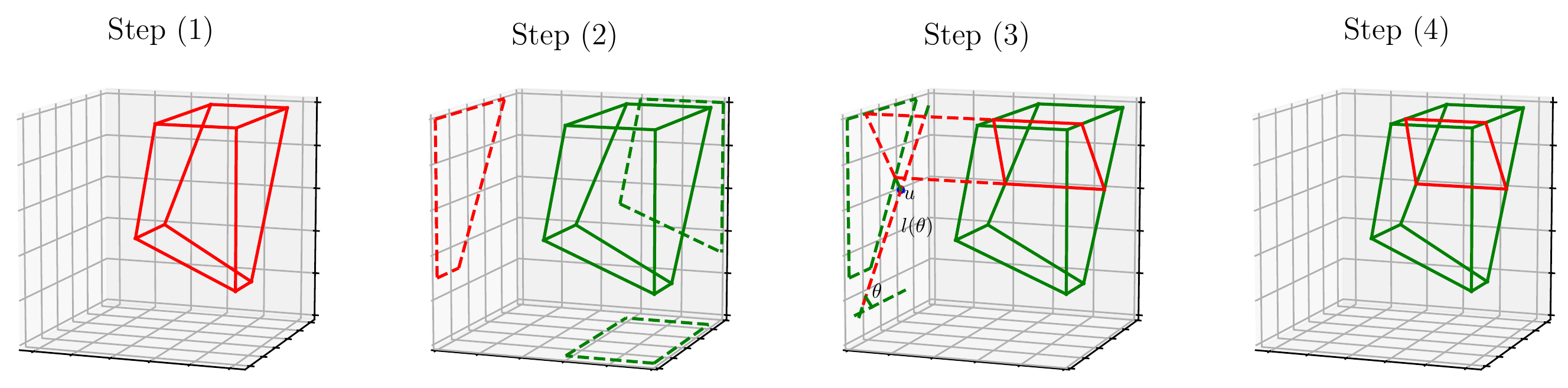}
\caption{Generating a cutting hyperplane in one branch of an example $3$-dimensional BSP-tree whose root node is a unit cube. We use red-line constituted polyhedrons (or polygons) to denote the sampled ones; the three dashed polygons in Step (2) denote all the two-dimensional projections of the polygon sampled in Step (1). The generations of $\theta$ and $\pmb{u}$ on a two-dimensional projection (green dashed polygon) in Step (3) follows the same way as in \cite{pmlr-v84-fan18b}.}
\label{fig:generative_process}
\end{figure*}

\section{Extending the BSP-tree process to $d$-dimensional space}


One motivation for extension of the BSP-tree process to multi-dimensional space is regression. Here, we might have $N$ labelled datapoints $\{(\pmb{x}_i, y_i)\}_{i=1}^N\in\mathbb{R}^{d}\times \mathbb{R}$, where we aim to predict the unknown labels $y_i\in\mathbb{R}$ from the $d$-dimensional predictors $\pmb{x}_i\in\mathbb{R}^d$. Other applications, such as classification, simply require straightforward changes to the likelihood constructions for the labels.


The proposed multidimensional BSP-Tree process works similarly as its original version. It is still a continuous-time Markov jump process where, for  ${\tau}_{l+1}>{\tau}_l>0$, the value taken at  ${\tau}_{l+1}$, which is denoted as $\boxplus_{\tau_{l+1}}$, is a BSP-Tree partition in a convex polyhedron $\square \subset \mathbb{R}^d$ and also a further refinement of the value taken at time $\tau_l$~(see Figure~\ref{fig:generativeprocesspartition}). For the domain (root node) $\square\subset\mathbb{R}^d$, the partition result $\boxplus_{\tau_l}$ is composed of a set of convex polyhedrons $\big\{\{\square_k\}_{k\in\mathbb{N}^+}:\cup_k \square_{k}=\square; \square_{k'}\cap \square_{k''}=\emptyset, \forall k'\neq k''\big\}$ and is recursively generated through a series of cutting hyperplanes. 

In order to extend the domain of the BSP-Tree process to $d$-dimensional ($d>2$) space while still keeping its self-consistency, we consider a reduced generative process where each cutting hyperplane is only allowed to have two degrees (dimensions) of freedom. In this way, each potential cutting hyperplanes on $\square_k$ is assumed to be parallel to the rest $d-2$ dimensions, except the selected two. 

Given the current partition $\boxplus_{\tau_{l-1}} = \{\square^{(k)}_{\tau_{l-1}}\}_{k=1}^l$ and $\tau$, the next cutting hyperplane is generated in the following steps (see the illustrations for Steps 1--4 in Figure~\ref{fig:generative_process}):
\begin{description} \label{partitiondescription}
    \item[(1)] Sample a candidate polyhedron (leaf node) $\square
    ^{(*)}$ from all the existing leaf nodes $\{\square^{(k)}_{\tau_{l-1}}\}_{k=1}^l$ in proportion to $\left\{\sum_{d_1,d_2} PE(\Pi_{d_1,d_2}(\square^{(k)}_{\tau_{l-1}}))\right\}_{k=1}^l$, where $(d_1, d_2)$ denotes an arbitrary pair of dimensions from the $d$ dimensions, $\Pi_{d_1, d_2}(\square)$ denotes the projection of $\square$ onto the dimensions of $(d_1,d_2)$, and $PE(\Pi_{d_1, d_2}(\square))$ denotes the perimeter of the projection (i.e., a 2-dimensional polygon);
    
    \item[(2)] Sample a pair of free dimensions $(d_1^{(*)}, d_2^{(*)})$ from all $1/2\cdot d(d-1)$ possible pairs in proportion to $\left\{PE(\Pi_{d_1,d_2}(\square^{(*)}))\right\}_{(d_1,d_2)}$;
    
    \item[(3)] On the projection $\Pi_{d_1^{(*)}, d_2^{(*)}}(\square^{(*)})$, sample a direction $\theta$ from $(0, \pi]$, where the probability density function is in proportion to the length of the line segment $\pmb{l}(\theta)$, onto which $\Pi_{d_1^{(*)}, d_2^{(*)}}(\square^{(*)})$ is projected in the direction of $\theta$; and sample the cutting position $\pmb{u}$ uniformly on the line segment $\pmb{l}(\theta)$. The proposed cutting hyperplane is formed as the straight line passing through $\pmb{u}$ and crossing through the projection $\Pi_{d_1^{(*)}, d_2^{(*)}}(\square^{(*)})$, orthogonal to $\pmb{l}(\theta)$ in the dimensions of $(d_1^{(*)},d_2^{(*)})$ and parallel to the rest $d-2$ dimensions\footnote{Generating a cutting line parameterized by $\theta$ and $\pmb{u}$ on the projection $\Pi_{d_1^{(*)}, d_2^{(*)}}(\square^{(*)})$ (two-dimensional polygon) follows the same sampling method used in the BSP-Tree process~\cite{pmlr-v84-fan18b}.};

    \item[(4)] Sample the incremental time for the new cut as $(\tau_l-\tau_{l-1}) \sim \textrm{Exp}\left(\sum_{k=1}^l\sum_{(i,j)\in\mathcal{D}}PE(\Pi_{d_i,d_j}(\square_{\tau_{l-1}}^{(k)}))\right)$. If $\tau_{l}>\tau$, reject the proposed cutting hyperplane and return $\{\square^{(k)}_{\tau_{l-1}}\}_{k=1}^l$ as the final partition structure; otherwise accept the proposed cutting hyperplane, increase $l$ to $l+1$ and go back to Step (1).
\end{description}
Sampling a two-dimensional pair (Step (2)) is the novel key step that helps extend the BSP-tree process to $d$-dimensional spaces; all other steps are the natural and logical extensions of the generative process of the two-dimensional BSP-tree process.

Through the above generative process, the cutting hyperplane can be parameterised as $H(k, (d_1, d_2), \theta, \pmb{u})=\{\pmb{x}\in\square^{(k^*)}|([x_{d_1}, x_{d_2}]-\pmb{u})(1;\tan\theta)^{\top}=0\}$, where $k^*$ denotes the index of the selected polygon (leaf node), $x_{d_1}$ denotes the $d_1$-th element of vector $\pmb{x}$, and $\pmb{u}$ is a two-dimensional vector denoting the position on the dimensions of $(d_1, d_2)$. The cutting hyperplane is parallel to all dimensions except $d_1$ and $d_2$ such that it is fully characterised on $(d_1, d_2)$. 

Moreover, the cutting hyperplane is only meaningful (i.e.~it intersects with $\square^{(k^*)}$) if and only if it intersects with the projection $\Pi_{d_1,d_2}(\square^{(k^*)})$ on dimensions $(d_1, d_2)$. As $\Pi_{d_1,d_2}(\square^{(k^*)})$ is a convex polygon, from Proposition 2 of~\cite{pmlr-v84-fan18b}, the measure of the hyperplane cuts on  dimensions $(d_1, d_2)$ 
{is uniform over}
the perimeter of $\Pi_{d_1,d_2}(\square^{(k^*)})$. Thus, the measure of the hyperplane cuts for $\square^{(k^*)}$ 
{is uniform over}
 the sum $|\mathcal{D}|$ of the perimeters of $\Pi_{d_1,d_2}(\square^{(k^*)})$ on all unique pairs of dimensions
i.e.~$\sum_{(i,j)\in\mathcal{D}} PE(\Pi_{d_i,d_j}(\square^{(k^*)}))$. 

When $d=2$, the above extended BSP-tree process reduces to the original two-dimensional version of~\cite{pmlr-v84-fan18b}. According to Propositions 1 and 2 in \cite{pmlr-v84-fan18b}, the likelihood of the cutting hyperplane on the two dimensions $(d_1^*, d_2^*)$ for $\square^{(k^*)}$ is $p(\theta, \pmb{u}|d_1^*,d_2^*,\square^{(k^*)})=\frac{1}{PE(\Pi_{d^*_1,d^*_2}(\square^{(k^*)}))}$. Extending this hyperplane to $d$-dimensions, the likelihood is 
$p(cut)=\frac{\sum_{(i,j)\in\mathcal{D}} PE(\Pi_{d_i,d_j}(\square^{(k^*)}))}{\sum_{k=1}^l\sum_{(i,j)\in\mathcal{D}} PE(\Pi_{d_i,d_j}(\square^{(k)}))} \cdot \frac{PE(\Pi_{d^*_1,d^*_2}(\square^{(k^*)}))}{\sum_{(i,j)\in\mathcal{D}} PE(\Pi_{d_i,d_j}(\square^{(k^*)}))} \cdot \frac{1}{PE(\Pi_{d^*_1,d^*_2}(\square^{(k^*)}))} = \frac{1}{\sum_{k=1}^l\sum_{(i,j)\in\mathcal{D}} PE(\Pi_{d_i,d_j}(\square^{(k)}))}$. 
That is, all potential partitions are equally favoured and the hyperplane cut is uniformly distributed, without involving additional (prior) knowledge about the cutting hyperplane. 

\paragraph{Self-consistency of the extended BSP-tree process:}

As a result of the particular technique of generating $d$-dimensional cutting hyperplanes, the BSP-tree process retains the self-consistency property in $d$-dimensions.

\begin{proposition}
The extended BSP-tree process is self-consistent in  $d$-dimensional ($d\geq 2$) space, and maintains distributional invariance when restricting its domain from a convex polyhedron to a sub-domain.
\end{proposition}
That is, if the extended BSP-Tree process on a finite convex polyhedron $\square$ is then restricted to a sub-region $\triangle, \triangle\subseteq\square$, then the resulting partitions restricted to $\triangle$ are distributed as if they were directly generated on $\triangle$ through the BSP-tree generative process. Subsequent application of the Kolmogorov Extension Theorem~\cite{chung2001course} states that the self-consistency property of the BSP-tree process then enables it to be defined on infinite multidimensional space.


This useful property can strongly simplify the sampling procedure and improve algorithmic efficiency in $d$-dimensional ($d>2$) implementations, which would otherwise be extremely complicated to implement (see inference sections, below).

\section{Binary space partitioning~(BSP)-forests}


We consider the label-predicting regression task outlined earlier, so that for observed data  $\{(\pmb{x}_i, y_i)\}_{i=1}^N\in\mathbb{R}^{d}\times \mathbb{R}$ we aim to predict the unknown labels $y_i\in\mathbb{R}$ from the $d$-dimensional predictors $\pmb{x}_i\in\mathbb{R}^d$. 

\subsection{Model}

We apply the extended BSP-tree process in a random forest style model. For the regression we attach {mean intensity parameters, $\mu$,} to the leaf nodes of each tree structure, and use the sum of the intensities from $m$ BSP-trees to predicting the unknown label. That is,
\begin{align} \label{eq:sum_of_tree}
y=\sum_{j=1}^mg\left(\pmb{x};\boxplus^{(j)}, \pmb{\mu}^{(j)}\right)+\epsilon, \quad \epsilon\sim\mathcal{N}(0, \sigma^2),
\end{align}
where $\pmb{\mu}^{(j)}=\{\mu^{(j)}_1, \cdots, \mu^{(j)}_{k^{(j)}}\}$ denotes the set of mean parameters associated with the leaf nodes of the $j$-th BSP-tree, $k^{(j)}$ is the number of leaf nodes in the $j$-th BSP-tree, $g\left(\pmb{x}; \boxplus^{(j)}, \pmb{\mu}^{(j)}\right)=\mu^{(j)}_{k_0}$ if $\pmb{x}$ belongs to the $k_0$-th leaf node of the $j$-th BSP-tree, and  $\sigma^2$ denotes the observation variance.



Determination of node assignments for $\pmb{x}$ follows the typical procedure used for a decision tree. To  begin, $\pmb{x}$ belongs to the root node of the BSP-tree $\boxplus^{(j)}$. On any non-leaf node, the cutting hyperplane $H(k, (d_1, d_2), \theta, \pmb{u})$ would then be used to determine the child node that  $\pmb{x}$ belongs to: $\{(x_{d_1}, x_{d_2})-\pmb{u})(1;\tan\theta)^{\top}<0\}$ or $\{(x_{d_1}, x_{d_2})-\pmb{u})(1;\tan\theta)^{\top}>0\}$. Through a sequence of such binary node assignments from the top to the bottom of the tree, $\pmb{x}$ would finally fall into a single leaf node $k_0$ of $\boxplus^{(j)}$, with $g$ then allocating the intensity $\mu^{(j)}_{k_0}$ to the regression.

In common with other regression forests, 
each 
individual tree only contributes proportionally $\frac{1}{m}$ of the model for label $y$, which prevents any one of the trees dominating the prediction. 
Further, because the ensemble comprises multiple BST-trees, each potentially selecting different dimensional pairs to describe the data in local regions, the BSP-forest is able to capture all pair-wise dependence in $d$-dimensions.
%
Similarly, the potentially different depths of each tree allows the BSP-forest to reflect different granularities of interactions between dimensions.

Each regression tree in the BSP-forest presents a hierarchical partition structure on $d$-dimensional space.
From this perspective, the {\it sum-of-trees} forest mechanism provides a compositional ``smoothing'' of the partition structure for local regions of the space, thereby greatly increasing the flexibility of the regression model. Given $m$, a BSP-forest is determined by $\{(\boxplus^{(j)}, \pmb{\mu}^{(j)})\}_{j=1}^m$, which includes the BSP-tree structures and their associated leaf node mean parameters, and the observational variance parameter $\sigma^2$.

\paragraph{Prior specification for $\sigma^2, \pmb{\mu}, m$:} 

We specify the prior distributions for $\sigma^2$, the number of trees in the BSP-forest $m$, and the mean parameters $\{\pmb{\mu}^{(j)}\}_{j=1}^m$ for all $m$ trees $\{\boxplus^{(j)}\}_{j=1}^{m}$ following~\cite{chipman2010bart}. Namely: (1) $\sigma^2\sim\textrm{IGamma}(1.5,\lambda)$ where, given an estimate of the sample variance $\hat{\sigma}^2$, $\lambda$ is set to satisfy the condition $F(\Hat{\sigma}^2;1.5, \lambda)=0.9$, where $F(\cdot)$ is the Inverse-Gamma c.d.f.; (2) As the labels $y$ are typically standardised, the prior for each mean parameter is specified as $\mu\sim N\left(0,\frac{1}{2\sqrt{m}}\right)$; (3) While a prior for $m$ can be specified, and inference performed by a reversible-jump type algorithm \cite{sisson05}, for simplicity we fix $m=50$ which we find performs well in practice.

\paragraph{Comparison with BART and MF:} 

The BSP-forest is directly related to two popular Bayesian regression-tree models: Bayesian additive regression trees~(BART) and the Mondrian forest (MF). The key differences and advantage here is that the BSP-forest uses {both angled cuts, and} hyperplane cuts constructed from two dimensions, rather than the one-dimensional, {axis-aligned} cuts in both BART and MF. 
This clearly provides a more flexible way to model the observed data, and as a result, lower tree depth is required to obtain the same or even improved modelling capability. 

\subsection{Inference}

Posterior inference for the BSP-forest is implemented using Markov chain Monte Carlo (MCMC). The joint distribution of all parameters, including the partition structures of $m$ extended BSP-trees $\{\boxplus^{(j)}\}_{j=1}^m$, the  leaf-node mean parameters of each tree $\{\pmb{\mu}^{(j)}\}_{j=1}^m$, and the observational variance  $\sigma^2$ can be written as:
\begin{align}
    p(\{\boxplus^{(j)}, \pmb{\mu}^{(j)}\}_j, \sigma^2, Y|\pmb{X}, \lambda) = p(Y|\{\boxplus^{(j)}, \pmb{\mu}^{(j)}\}_j, \pmb{X}, \sigma^2)\nonumber \\
    \cdot \prod_{j=1}^m\left[p(\boxplus^{(j)}|\lambda) p(\pmb{\mu}^{(j)})\right] p(\sigma^2), \nonumber
\end{align}
where $Y=\{y_1,\ldots,y_N\}$ and $\pmb{X}=\{\pmb{x}_1,\ldots,\pmb{x}_N\}$.
Algorithm~\ref{algo:BSP_Forest} outlines the MCMC algorithm, which iteratively samples the variance $\sigma^2$~(Line 3) and each BSP-tree structure $\boxplus^{(j)}$ with mean parameters $\pmb{\mu}^{(j)}$)~(Line 5).

\begin{algorithm}[t]
\caption{MCMC for inferring BSP-Forest}
\label{algo:BSP_Forest}
\begin{algorithmic}[1]
\REQUIRE data set $\pmb{X}$, label set $Y$, number of iterations $T$
\ENSURE samples of $\{\boxplus^{(j)}, \pmb{\mu}^{(j)}\}_{j=1}^m$ and $\sigma^2$
\STATE initialize $m$ BSP-trees $\{\boxplus^{(j)}, \pmb{\mu}^{(j)}\}_{j=1}^m$
\FOR{$t=1, \cdots, T$}
\STATE sample $\sigma^2$ according to Eq.~(\ref{eq:sigma_sampling});
\FOR{$j=1, \cdots, m$}
\STATE implement C-SMC samplers for $\boxplus^{(j)}$ and $\pmb{\mu}^{(j)}$ using Algorithm~\ref{algo:C_SMC};
\ENDFOR
\ENDFOR
\end{algorithmic}
\end{algorithm}

\paragraph{Sampling $\sigma^2$:} 
Through conjugacy, the full posterior conditional distribution of $\sigma^2$ is
\begin{align} \label{eq:sigma_sampling}
    \sigma^2\sim \textrm{IGamma}\left(\frac{3+N}{2}, \hat{\lambda}+\frac{Ess}{2}\right)
\end{align}
where $Ess=\sum_{i=1}^N\left(y_i-\sum_{j=1}^mg(\pmb{x}_i;\boxplus^{(j)}, \pmb{\mu}^{(j)})\right)^2$.

\begin{algorithm*}[t]
\caption{The C-SMC sampler for the $j$-th BSP-tree $\boxplus^{(j)}$ at the $t$-th iteration} 
\label{algo:C_SMC}
\begin{algorithmic}[1]
\REQUIRE $\pmb{X}, Y, \sigma^2$, number of particles $N$, segments of the time line $\{(\tau_{s}, \tau_{s+1})\}_{s=0}^S$, $\boxplus^{(j)}$ and $\pmb{\mu}^{(j)}$ obtained at the last iteration, weights of the last selected particle over $S$ segments $\omega(\boxplus_{1:S}^{(j)})$
\ENSURE $\boxplus^{(j)}(n^*), \pmb{\mu}^{(j)}(n^*)$,  $\omega(\boxplus_{1:S}^{(j)}(n^*))$
\STATE for $s=0$, initialize $N-1$ particles $\boxplus^{(j)}_s(2:N)=\square$, means $\pmb{\mu}_{s}^{(j)}(2:N)=0$, costs $c^{(j)}_s(2:N)=0$; 
\FOR{$s=1, \cdots, S$}
    \STATE for $n=2, \cdots, N$, if $\tau_{s}<\sum_{s'=0}^{s-1}c^{(j)}_{s'}(n)<\tau_{s+1}$, recursively sample $\{H_{i}^{(j)}(n), \hat{c}_{i}^{(j)}(n)\}_{i=1}^V$ using Algorithm \ref{algo:cut_generation} and sample $\{\pmb{\mu}_{i}^{(j)}(n)\}_{i=1}^V$ from its posterior until $\sum_{s'=0}^{s}c^{(j)}_{s'}(n)>\tau_{s+1}$; otherwise set $H_s^{(j)}(n)=H_{s-1}^{(j)}(n)$, $\pmb{\mu}_s^{(j)}(n)=\pmb{\mu}_{s-1}^{(j)}(n)$,  $c_s^{(j)}(n)=0$;\\
    \% Notes: $H_s^{(j)}(n)=\{H_i^{(j)}(n)\}_{i=1}^V$, $\pmb{\mu}_s^{(j)}(n)=\{\pmb{\mu}_i^{(j)}(n)\}_{i=1}^V$, $c_s^{(j)}(n)=\sum_{i=1}^V \hat{c}_i^{(j)}(n)$ and $V$ denotes the number of cuts falling in $(\tau_{s}, \tau_{s+1}]$
    \STATE for $n=2, \cdots, N$, compute weight $\omega(\boxplus_s^{(j)}(n)):=\frac{\textrm{prior}(\pmb{\mu}^{(j)}_s(n))p(Y|\pmb{X}, H_{1:s}^{(j)}(n), \pmb{\mu}_{1:s}^{(j)}, \pmb{\mu}^{(-j)}, \sigma^2)}{\textrm{posterior}(\pmb{\mu}^{(j)}_s(n))p(Y|\pmb{X}, H_{1:(s-1)}^{(j)}(n), \pmb{\mu}_{1:(s-1)}^{(j)}, \pmb{\mu}^{(-j)}, \sigma^2)}$;\\ 
    \% Notes: $\pmb{\mu}^{(-j)}$ denotes the sum of mean parameters on $\pmb{X}$ from all the other BSP-trees except $\boxplus^{(j)}$
    \STATE for $n=1, \cdots, N$, normalize weights $W_{s}^{(j)}(n): = \frac{\omega(\boxplus_{s}^{(j)}(n))}{\sum_{n=1}^N\omega(\boxplus_{s}^{(j)}(n))}$;
    \STATE if $s<S$, for $n=2, \cdots, N$, sample $A_{s}(n) \sim \textrm{Discrete}(W_{s}^{(j)}(1:N))$ and set $H^{(j)}_{1:s}(n)=H^{(j)}_{1:s}(A_{s}(n))$,  $\pmb{\mu}^{(j)}_{1:s}(n)=\pmb{\mu}^{(j)}_{1:s}(A_{s}(n))$,  $c^{(j)}_{1:s}(n)=c^{(j)}_{1:s}(A_{s}(n))$; otherwise sample $n^*\sim \textrm{Discrete}(W_{S}^{(j)}(1:N))$;
\ENDFOR
\end{algorithmic}
\end{algorithm*}
\begin{algorithm*}[t]
\caption{Implementing a cutting hyperplane on the convex hull of one of the leaf nodes in $\boxplus^{(j)}$} 
\label{algo:cut_generation}
\begin{algorithmic}[1]
\REQUIRE $\pmb{X}, Y, \sigma^2$, polyhedrons (current leaf nodes) $\{\square^{(k)}\}_{k=1}^K$ in $\boxplus^{(j)}$
\ENSURE cutting hyperplane $H\left(k^*, (d^*_1, d^*_2), \theta, \pmb{u}\right)$ and cost $c$
\FOR{$(d_1, d_2) \in \{(1,2), \cdots, (1,d), (2,3), \cdots, (d-1,d)\}, k=1,\cdots, K$ }
    \STATE project $\{\pmb{x}_i\}_{i:\pmb{x}_i\in\square^{(k)}}$ onto the dimensions of $(d_1, d_2)$ to get $\{x_{i,d_1}, x_{i,d_2}\}_{i:\pmb{x}_i\in\square^{(k)}}$;
    \STATE calculate the convex hull on $\{x_{i,d_1}, x_{i,d_2}\}_{i:\pmb{x}_i\in\square^{(k)}}$, denoted by $\triangle^{(k)}_{(d_1,d_2)}$; 
    \STATE calculate the perimeter of $\triangle^{(k)}_{(d_1,d_2)}$, denoted by $PE(\triangle^{(k)}_{(d_1,d_2)})$;
\ENDFOR
\STATE sample $k^*,(d^*_1,d^*_2)$ in proportion to $\left\{PE(\triangle^{(1)}_{(1,1)}),\cdots,PE(\triangle^{(k)}_{(1,d)}),PE(\triangle^{(k)}_{(2,3)}),\cdots,PE(\triangle^{(K)}_{(d-1,d)})\right\}$; 
\STATE sample $\theta, \pmb{u}$ on the projection of the convex hull $\triangle^{(k^*)}_{(d^*_1,d^*_2)}$;
\STATE sample the cost  $c\sim\textrm{Exp}\left(\sum_k\sum_{d_1,d_2}PE(\triangle^{(k)}_{(d_1,d_2)})\right)$;
\end{algorithmic}
\end{algorithm*}

\paragraph{Sampling $\{\boxplus^{(j)}, \pmb{\mu}^{(j)}\}_{j=1}^{m}$} We use the Conditional-Sequential Monte Carlo~(C-SMC) sampler~\cite{andrieu2010particle} to infer the partition structures of all the $m$ extended BSP-trees $\{\boxplus^{(j)}\}_{j=1}^{m}$ and the mean parameters $\{\pmb{\mu}^{(j)}\}_{j=1}^m$. There is a key difference between our sampler and that used in~\cite{LakRoyTeh2014a, pmlr-v84-fan18b}: These previous works take node splitting events on the time line as the dimensions of the sampled variable. The particles at the same step are under different budget constraints and the target distribution of the C-SMC samplers may not be well mixed. In our implementation, we adopt fixed intervals $\{(\tau_{s}, \tau_{s+1})\}_{s=0}^{S}$ ($\tau_0=0, \tau_{S+1}=\tau$) on the time line as the dimensions of the variable sampled by the C-SMC sampler. $\forall$ step $s$, all the particles are under the same budget constraints $(0, \tau_s]$ and each dimension might involve one, more than one, or even no splitting events. This helps to fix the number of ``dimensions'' in C-SMC and improve the efficiency of the sampler. Algorithm~\ref{algo:C_SMC} shows the detail strategy to infer the tree structure in the $(t+1)$-th iteration for the $j$-th BSP-tree. 

\begin{figure}[t]
\centering
\includegraphics[width =  0.2 \textwidth]{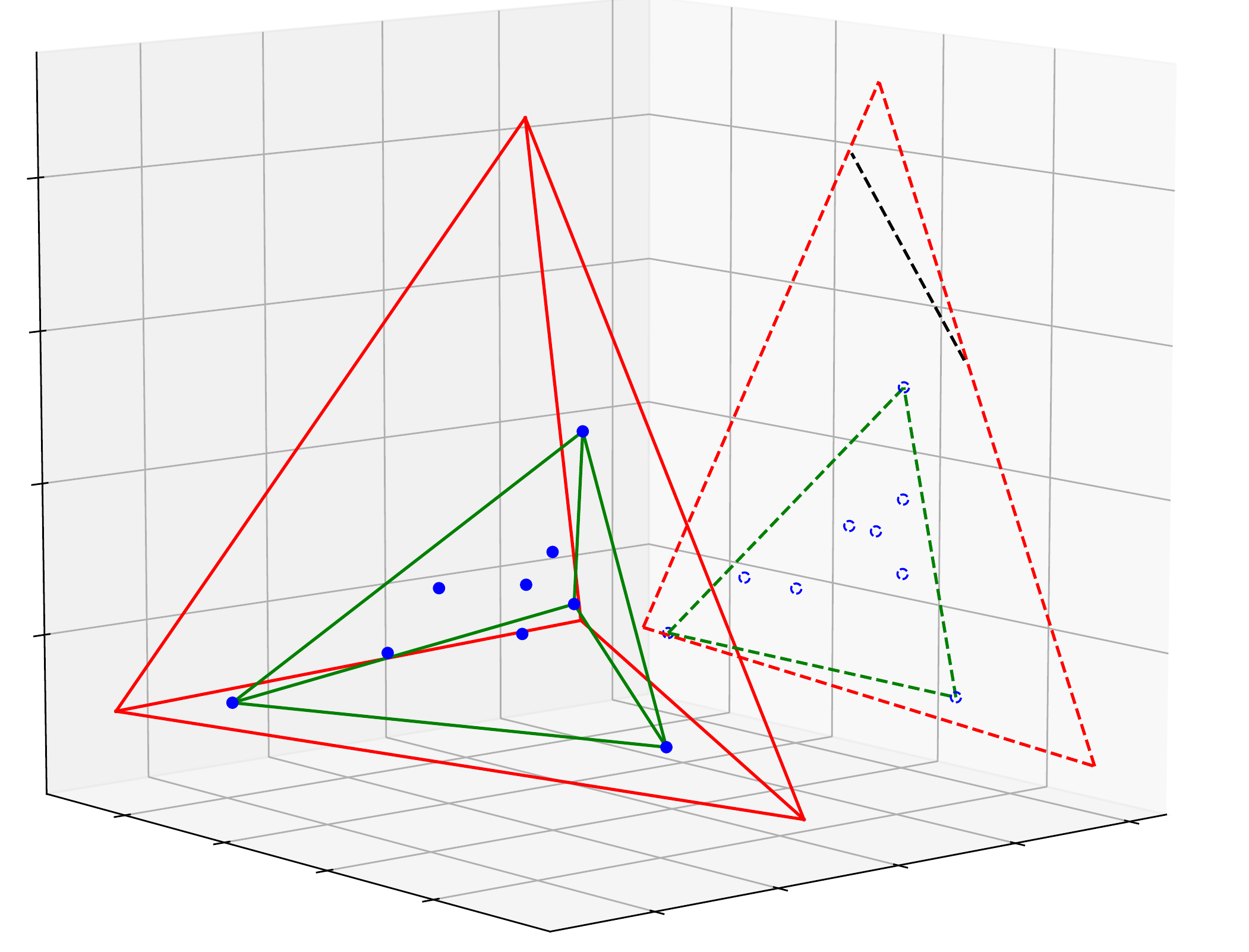}
\caption{Visualization of calculating projection of convex hull in the example of $3$-dimensional space. Polyhedron with red solid lines denotes the ``full'' block generated by the cutting hyperplanes; polygon with red dashed lines denotes the polyhedron's projection in two dimensions; blue points denotes the data $\pmb{x}$ belong to this block; blue dashed circles denotes $\pmb{x}$'s projection in two dimensions; polyhedron with green solid lines denotes the convex hull spanned by $\pmb{x}$ and polygon with green dashed lines denotes the convex hull's projection in two dimensions. The calculation of the projections (polygons with dashed lines) requires the vertices of the polyhedron and the cut (black dashed line) outside the convex hull's projection does not change the model's prediction on labels.}
\label{fig:convex_hull}
\end{figure}

\paragraph{Implementing cutting hyperplane on convex hull:} Cutting the full $d$-dimensional polyhedron generated from a series of cutting hyperplanes on the original domain, is both a challenging and unnecessary task. This is because:  (1) complete indexing of this polyhedron requires extensive geometric calculations, including calculating the intersection of multiple hyperplanes, deciding if two hyperplanes intersect {in the existence of } other hyperplanes, and listing all the vertices of the polyhedron; (2) in some cases, the cutting hyperplane occurs outside the convex hull~(i.e.~the minimum convex polyhedron that contains all the points in this block), it does not partition the available data and thereby influence the regression. In practice, while data usually lies in a small portion of the feature space, implementing cutting hyperplanes directly on this polyhedron is particularly inefficient.

To address these issues, we implement the hyperplane cut on the convex hull only. The projection of the convex hull on any two dimensions can be obtained by simply slicing the $\pmb{x}$ coordinates on these two dimensions and then using a conventional convex hull detection algorithm~(such as the Graham Scan algorithm~\cite{Graham1972}. Algorithm~\ref{algo:cut_generation} describes our approach to generate the hyperplane cut in the convex hull and Fig.~\ref{fig:convex_hull} visualises
the convex hull projection in $3$-dimensional space.

Due to the self-consistency property of the (extended) BSP-tree process, we conveniently have that: 
\begin{proposition}
The hyperplane restricted on the convex hull is distributed the same as if we first partition on the ``full'' polyhedron and then restrict attention to the convex hull. Both of these two methods lead to the idential  equilibrium distribution in the MCMC algorithms.
\end{proposition}

\paragraph{Computational cost}
The computational cost of the BSP-Forest is almost the same as the batched Mondrian Forest, except that the number of candidate features (used to generate a hyperplane) increases from $\mathcal{O}(d)$ to $\mathcal{O}(d^2)$.

\section{Related Work}

Random forests have been proposed for decades and the volume of literature may be too huge to be reviewed here. The interested readers can refer to~\cite{criminisi2012decision} for a recent review. For the structure of a tree in the forest, the node splitting in a binary tree usually consists of two steps: (1) choose the most suitable dimension to cut; (2) find the best cutting position on the chosen dimension. A widely adopted strategy is to optimize some information gain related criterion in a greedy way. In terms of the decision tree, a random forest uses the \emph{sum-of-trees} strategy to further improve the model capability. Some additional strategies to decorrelate the trees include bagging and subsampling the data set. 

{Here we use the Breimain-Random Forest~\cite{breiman2001random} and Gradient Boosting regressor~\cite{friedman1991} as two representative works. The Breimain-RF adopts the bagging strategy and chooses a subset of the $d$ dimensions for each tree. The Gradient Boosting regressor uses the boosting strategy, which is to propose regression trees to cater for the negative gradient of the loss function at the current step.} 

For the Bayesian decision trees, Bayesian Additive Regression Trees (BART)~\cite{chipman2010bart} and the Mondrian Forest~\cite{LakRoyTeh2014a} are two representative counterparts of the proposed BSP-Forest. BART assigns probability distributions to the structure variables of all the trees, while MF uses the Mondrian process as the prior of the tree structure for the forest. In contrast to these two methods which merely select one dimension to conduct node cut to generate the tree structure, the BSP-Forest adopts two dimensions together to form a hyperplane cut in the feature space.

Multivariate decision trees~\cite{brodley1995multivariate} (single-tree model) may be the closest one to the proposed BSP-tree in term of the inter-dimensional dependence. Its less popularity might be due to its high computational cost in finding an optimized cutting hyperplane. Compared to our BSP-tree, the lack of prior information in the multivariate decision tree requires more regularization techniques.

\section{Experiments}
\subsection{Toydata: Friedman's Function}
We first evaluate the performance of the BSP-Forest on the Friedman's function~\cite{friedman1991, chipman2010bart, linero2017bayesian}. In this test setting, each data point $\pmb{x}$ is generated from the uniform distribution, while its label $y$ takes the following form:
\begin{align} \label{eq:friedman_test}
    y=10\sin(\pi x_1x_2)+20\left(x_3-\frac{1}{2}\right)^2+10x_4+5x_5+\epsilon \nonumber
\end{align}
where $\epsilon\sim\mathcal{N}(0, \sigma^2), \sigma^2=1$ denotes the white noise effect and $x_d$ denotes the $d$-th dimension of $\pmb{x}$.
The Friedman's function consists of two nonlinear terms, two linear terms and an interaction term between the dimensions.

\begin{figure}[t]
\centering
\includegraphics[width =  0.4 \textwidth]{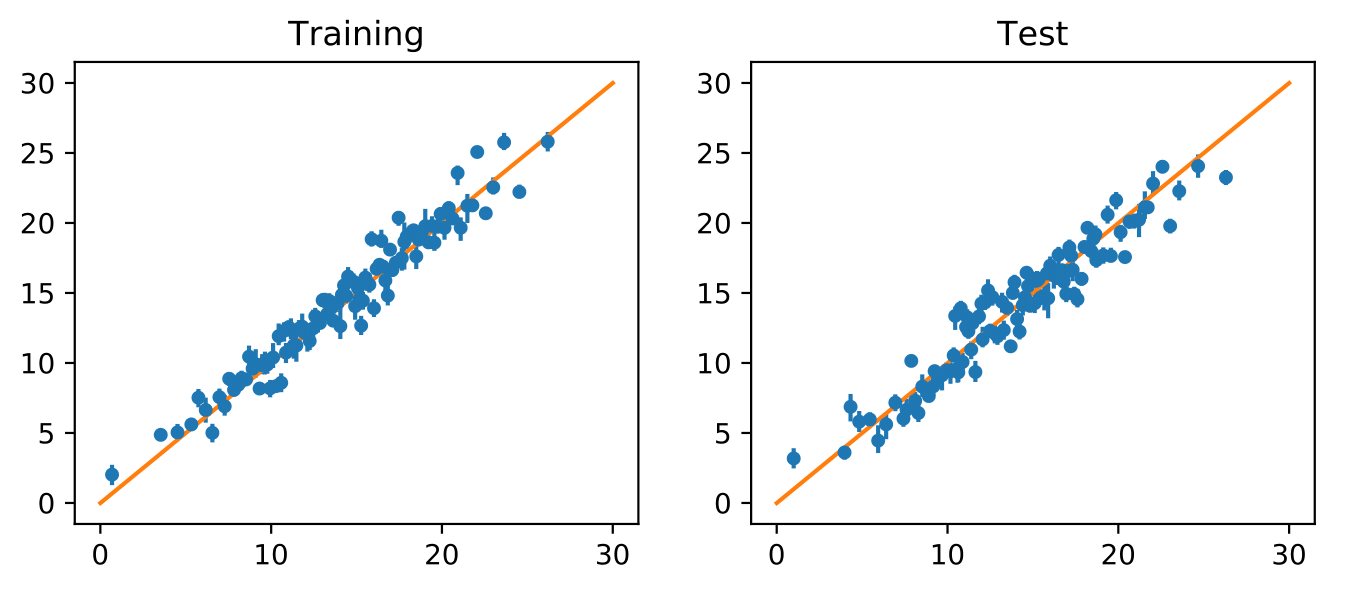}
\caption{Results on Friedman's function ($10$ dimensions). $x$-axis: ground-truth of $f(x)$; $y$-axis: predictive $\hat{f}(x)$.}
\label{fig:simple_result}
\end{figure}

\paragraph{Posterior Mean Estimation}
We begin with a simple application on posterior mean estimates of the BSP-Forest. The number of dimensions is set to $10$, such that $f(x)$ considers only the first $5$ dimensions of the data points and the rest five dimensions are irrelevant; the number of data points $N$ is set to $300$. The results in Figure~\ref{fig:simple_result} show the estimated $\hat{f}(x)$ against the groundtruth $f(x)$ on the training and test label sets. There is a vertical line on each point to indicate the $90\%$ posterior confidence interval. We can see that, on the training data set, the predicted values $\hat{f}(x)$ correlate very well with the true values $f(x)$, and the confidence intervals tend to cover the true values as well. On the test data set, one can observe a slight degradation of the correlation and wider intervals, which indicate a larger variance estimation of $f(x)$ at new $x$ values. 

\begin{figure}[t]
\centering
\includegraphics[width =  0.43 \textwidth]{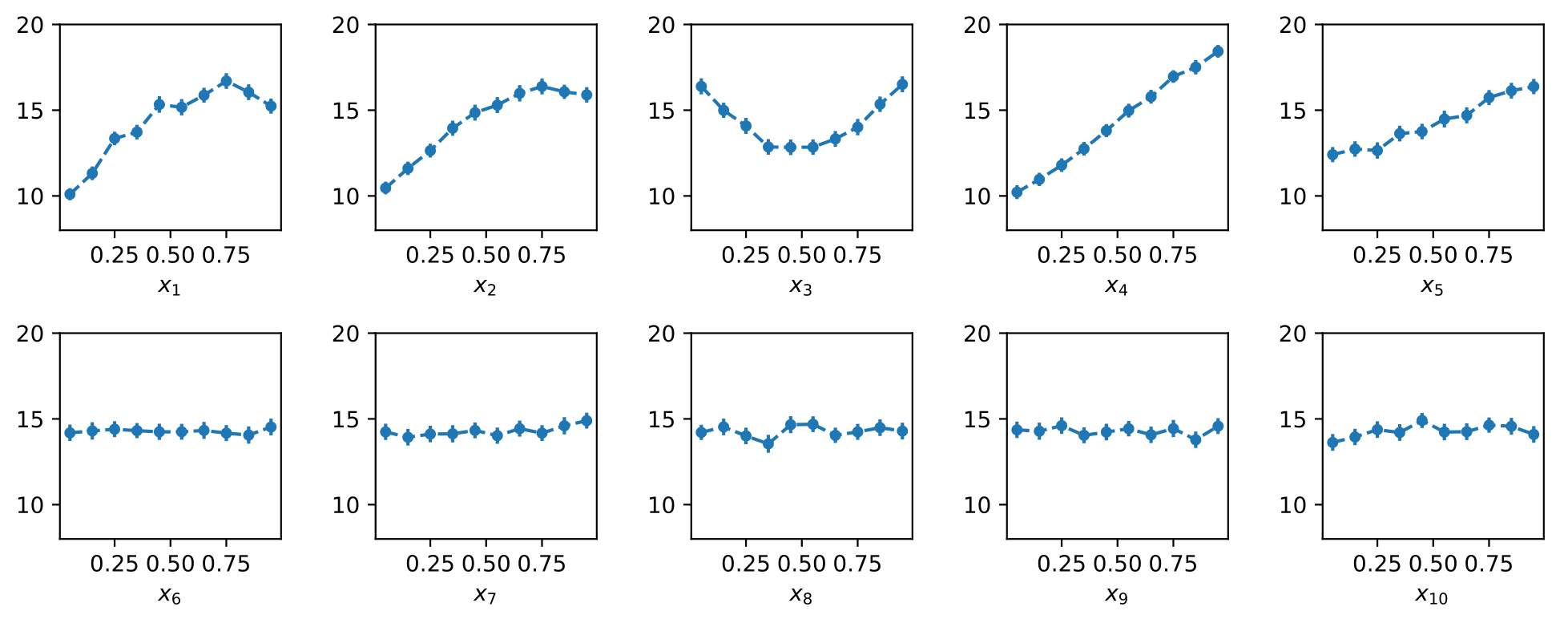}
\caption{Partial dependence plots for the $10$ dimensions on the Friedman's function data.}
\label{fig:partial_dependence}
\end{figure}

\paragraph{Partial Dependence}
Partial dependence is commonly used in Statistics to show the effect of adding features to a model. Figure~\ref{fig:partial_dependence} shows the plots of points and interval estimates of the partial dependence functions for $x_1, \cdots, x_{10}$ from the MCMC samplers. The nonzero marginal effects of $x_1, \cdots, x_5$ and the zero marginal effects of $x_6, \cdots, x_{10}$ seem to be completely consistent with the form of $f(x)$. 

\begin{table*}[t]
\centering
\caption{Test results on real-world datasets (RMAE$\pm$std)}
\label{table:RMAE_value}
\begin{tabular}{l|ccccc}
  \hline
  Data Sets  &{Random Forest} & {Gradient Boosting} & {BART} & {Mondrian Forest} & {BSP-Forest} \\
  \hline
Concrete     & $3.11\pm 0.23 $  & $4.54\pm 0.17 $  & $3.34\pm 0.41 $  & $3.18\pm 0.24 $  & $ \pmb{3.07\pm 0.08} $ \\
  diamonds     & $0.007\pm 0.002 $  & $0.011\pm 0.003 $  & $0.010\pm 0.004 $  & $0.013\pm 0.004 $  & $\pmb{0.005\pm 0.0003} $ \\
  hatco     & $0.28\pm0.03 $  & $0.26\pm 0.02 $ & $0.24\pm0.07 $  & $0.25\pm 0.03 $& $\pmb{0.21\pm0.05} $ \\
  servo     & $\pmb{0.23\pm 0.07} $  & $0.26\pm 0.05 $  & $0.27\pm 0.08 $  & $0.24\pm 0.04 $  & $\pmb{0.23\pm 0.04} $ \\
  tecator     & $3.27\pm 0.84 $  & $2.34\pm 0.74 $  & $2.84\pm 0.39 $  & $2.16\pm 0.46 $  & $\pmb{2.09\pm 0.61} $ \\
  \hline
\end{tabular}
\end{table*}

\begin{figure}[t]
\centering
\includegraphics[width =  0.4 \textwidth]{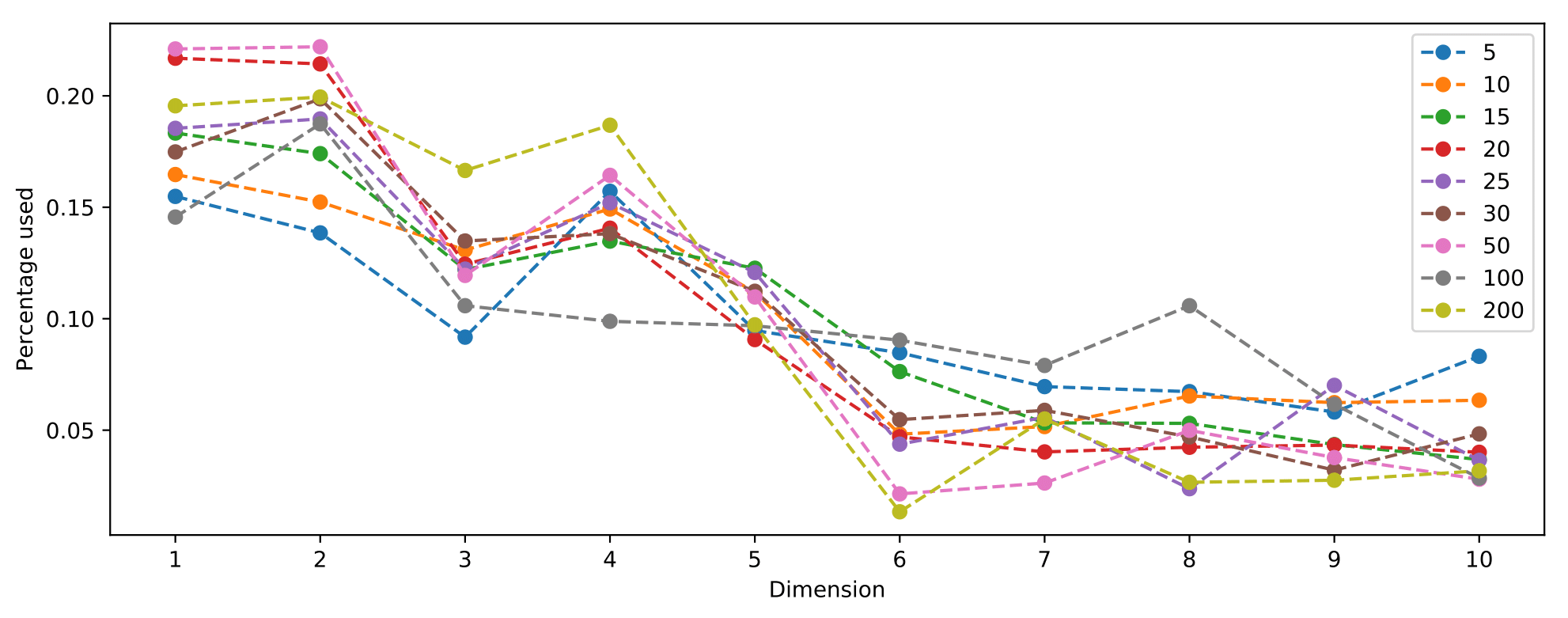}
\caption{Average usage of dimensions in terms of different number of trees in the BSP-Forest. Note that dimensions $1-5$ are the groundtruth ones that determine the label $y$.}
\label{fig:dim_use}
\end{figure}

\paragraph{Dimension Usage}
The BSP-Forest can also be used to identify the dependent dimensions that are most correlated to the function $f(x)$. This can be completed by recording the pair of dimensions used for each regression tree and count the times of involvements of these dimensions. Figure~\ref{fig:dim_use} plots the proportions of $x_1, \cdots, x_{10}$ for the number of trees $m=10, 20, 30, 50, 100$. As $m$ gets smaller, the fitted sum-of-trees models increasingly incorporate only those dimensions that are critical to explain the label $y$. Without making use of any assumptions or information about the actual function, the BSP-Forest has exactly identified the underlying dimensions that determine $f(x)$. 

\begin{figure}[t]
\centering
\includegraphics[width =  0.35 \textwidth]{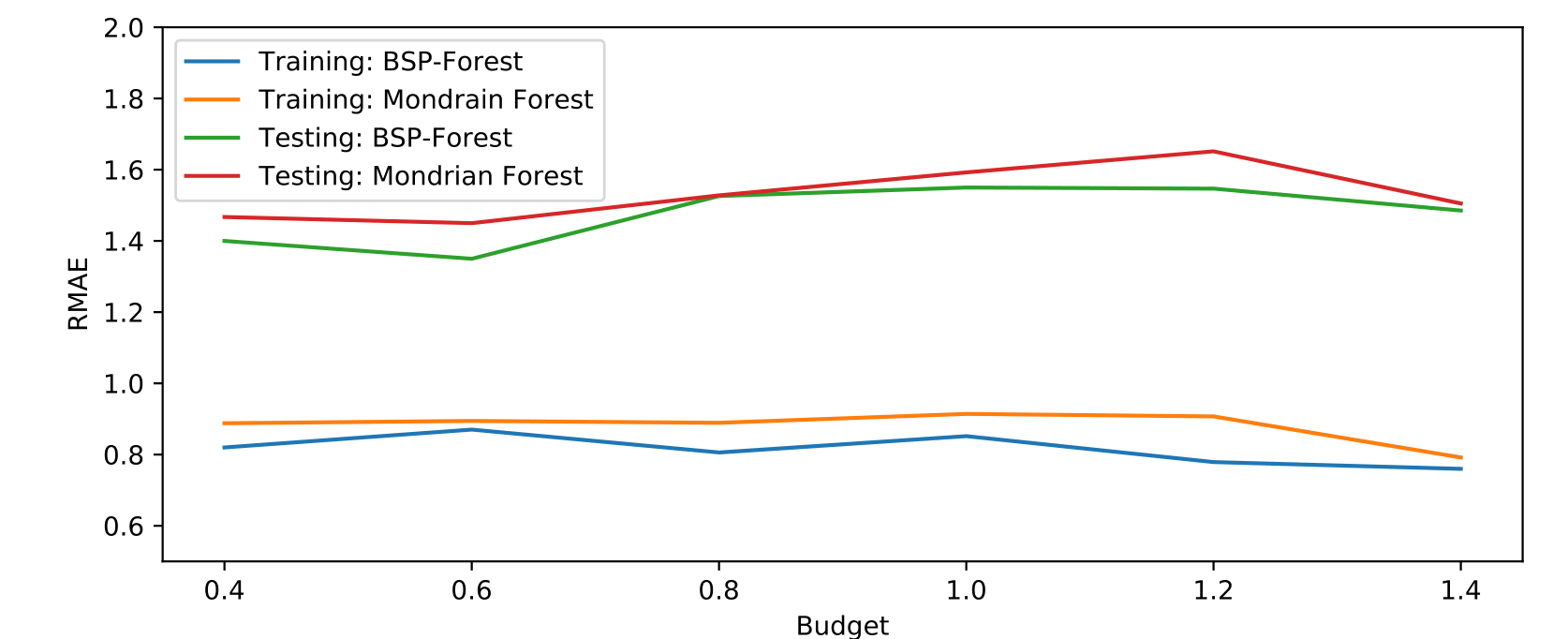}
\caption{RMAE performance on different budget values}
\label{fig:RMAE_budget}
\end{figure}

\begin{figure}[t]
\centering
\includegraphics[width =  0.4 \textwidth]{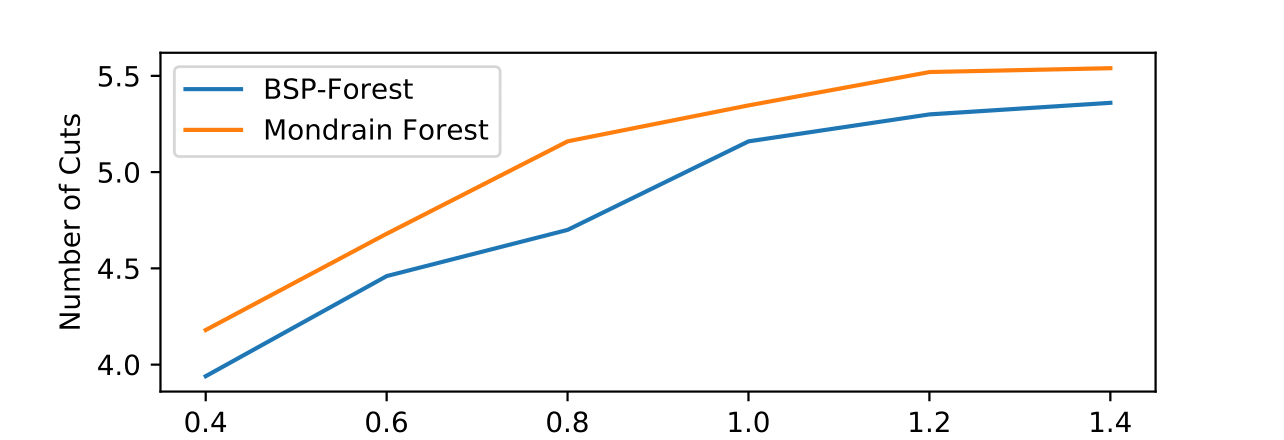}
\caption{Number of cuts on different budget values}
\label{fig:numcuts_budget}
\end{figure}

\paragraph{Performance Evaluation w.r.t. Budget}
For the Mondrian Forest and the BSP-Forest, the parameter determines the expected depth of the BSP-tree, the budget $\lambda$, has more impact on their performance. Figures~\ref{fig:RMAE_budget} and~\ref{fig:numcuts_budget} show the Root Mean Absolute Error~(RMAE) and the average number of cuts for the BSP-Forest and the Mondrian Forest, while budget is taking different values from $0.4$ to $1.4$. We can see that the BSP-Forest can obtain better RMAE performance on all budget values. Also, the number of cuts in the BSP-Forest is always smaller than that in the Mondrian Forest. The efficiency in the number of cuts is consistent with our expectation on the BSP-Forest. Although the curve seems quite flat across values of $\lambda$ considered, the reason may be that the average number of cuts does not change much: it only changes from 4 to 5 in Figure 8, as the budget varies from 0.4 to 1.4.


\begin{table}[t]
\centering
\caption{Real-world dataset information}
\label{table:dataset_information}
\begin{tabular}{l|cc|l|cc}
  \hline
  Name  & $d$ & $N$ & name & $d$ & $N$ \\
  \hline
  Concrete   & $9 $  & $1030$   &  diamonds & $9$  & $ 53940 $ \\
  hatco   & $14$  & $100$  &  servo & $4$  & $ 167 $ \\
  tecator   & $22$  & $215$   \\
  \hline
\end{tabular}
\end{table}

\subsection{Real-world Datasets}
The performance of the BSP-Forest is compared with other tree-based models on a number of real-world data sets. The comparison methods we include Bremain's Random Forest~\cite{breiman2001random}, Gradient Boosting regressor~\cite{friedman2001greedy}, Bayesian Additive Regression Trees (BART)~\cite{chipman2010bart}, batch version of the Mondrian Forest (MF)~\cite{NIPS2010_4151}. The implementations of Bremain's Random Forest and Gradient Boosting regressor are imported from the Python module of {\it Scikit-Learn}~\cite{scikit-learn}. BART is implemented through the {\it R} package of {\it BART}~\cite{BART_r_package}. We implement the Mondrian Forest through the same strategy as the BSP-Forest here. 

For a fair comparison, we set the number of trees in all the comparison methods as $50$. Except for this, the other parameters of Bremain's RF, Gradient Boosting regressor and BART are set as default in the modules. The budget of MF and the BSP-Forest is set to $\lambda =0.7$. In addition, we add scaling parameters to the distribution of the cost variable such that the expectation of the cost at stage $0$ is $\mathbb{E}(c)=0.25$ for both MF and BSP-Forest. 

Five real-world datasets~\cite{Lichman:2013} are used for the performance evaluation (see Table~\ref{table:dataset_information}). We  normalize the data $X\in\mathbb{R}^d$ into $[0, 1]^d$. Each data set is randomly divided into five equal parts where $4$ out of $5$ are used for training while the rest for testing. We report the Root Mean Absolute Error~(RMAE) as a robust measure over $10$ random runs on the datasets in Table~\ref{table:RMAE_value}. It is easily to see that our BSP-Forest performs the best among all the compared methods.

\section{Conclusion}
This paper makes the first endeavor to extend the BSP-Tree process to a $d$-dimensional ($d>2$) space. By designing a subtle strategy to only sample two free dimensions each time from the space, the extended BSP-Tree process can retain the essential self-consistency property. We further take the extended BSP-trees to construct the BSP-Forest for regression. Compared to other (Bayesian) regression forests, the BSP-Forest can perform best due to its flexibility.

\section{Acknowledgements}
Xuhui Fan and Scott A.~Sisson are supported by the Australian Research Council through the Australian Centre of Excellence in Mathematical and Statistical Frontiers (ACEMS, CE140100049), and Scott A.~Sisson through the Discovery Project Scheme (DP160102544). Bin Li is supported by Fudan University Startup Research Grant and Shanghai Municipal Science \& Technology Commission (16JC1420401).

\bibliography{StochasticPartitionProcessBase}

\begin{thebibliography}{10}

\bibitem{airoldi2009mixed}
Edoardo~M. Airoldi, David~M. Blei, Stephen~E. Fienberg, and Eric~P. Xing.
\newblock Mixed membership stochastic blockmodels.
\newblock In {\em NIPS}, pages 33--40, 2009.

\bibitem{andrieu2010particle}
Christophe Andrieu, Arnaud Doucet, and Roman Holenstein.
\newblock Particle markov chain monte carlo methods.
\newblock {\em {Journal of the Royal Statistical Society: Series B (Statistical
  Methodology)}}, 72(3):269--342, 2010.

\bibitem{breiman2001random}
Leo Breiman.
\newblock Random forests.
\newblock {\em Machine learning}, 45(1):5--32, 2001.

\bibitem{brodley1995multivariate}
Carla~E Brodley and Paul~E Utgoff.
\newblock Multivariate decision trees.
\newblock {\em Machine learning}, 19(1):45--77, 1995.

\bibitem{chipman2010bart}
Hugh~A. Chipman, Edward~I. George, and Robert~E. McCulloch.
\newblock Bart: Bayesian additive regression trees.
\newblock {\em The Annals of Applied Statistics}, 4(1):266--298, 2010.

\bibitem{criminisi2012decision}
Antonio Criminisi, Jamie Shotton, Ender Konukoglu, et~al.
\newblock Decision forests: A unified framework for classification, regression,
  density estimation, manifold learning and semi-supervised learning.
\newblock {\em Foundations and Trends{\textregistered} in Computer Graphics and
  Vision}, 7(2--3):81--227, 2012.

\bibitem{NIPS2018_RBP}
Xuhui Fan, Bin Li, and Scott Sisson.
\newblock Rectangular bounding process.
\newblock In {\em NeurIPS}, pages 7631--7641, 2018.

\bibitem{pmlr-v84-fan18b}
Xuhui Fan, Bin Li, and Scott~A. Sisson.
\newblock The binary space partitioning-tree process.
\newblock In {\em AISTATS}, volume~84 of {\em Proceedings of Machine Learning
  Research}, pages 1859--1867, 2018.

\bibitem{xuhui2016OstomachionProcess}
Xuhui Fan, Bin Li, Yi~Wang, Yang Wang, and Fang Chen.
\newblock {The Ostomachion Process}.
\newblock In {\em AAAI Conference on Artificial Intelligence}, pages
  1547--1553, 2016.

\bibitem{friedman1991}
Jerome~H. Friedman.
\newblock Multivariate adaptive regression splines.
\newblock {\em The Annals of Statistics}, 19:1--67, 1991.

\bibitem{friedman2001greedy}
Jerome~H Friedman.
\newblock Greedy function approximation: a gradient boosting machine.
\newblock {\em Annals of statistics}, pages 1189--1232, 2001.

\bibitem{Graham1972}
Ronald~L. Graham.
\newblock An efficient algorithm for determining the convex hull of a finite
  planar set.
\newblock {\em Inf. Process. Lett.}, 1(4):132--133, 1972.

\bibitem{karrer2011stochastic}
Brian Karrer and Mark~E.J. Newman.
\newblock Stochastic blockmodels and community structure in networks.
\newblock {\em Physical Review E}, 83(1):016107, 2011.

\bibitem{kemp2006learning}
Charles Kemp, Joshua~B. Tenenbaum, Thomas~L. Griffiths, Takeshi Yamada, and
  Naonori Ueda.
\newblock Learning systems of concepts with an infinite relational model.
\newblock In {\em AAAI}, volume~3, pages 381--388, 2006.

\bibitem{chung2001course}
Kai lai Chung.
\newblock {\em A Course in Probability Theory}.
\newblock Academic Press, 2001.

\bibitem{LakRoyTeh2014a}
Balaji Lakshminarayanan, Daniel~M. Roy, and Yee~Whye Teh.
\newblock {Mondrian} forests: Efficient online random forests.
\newblock In {\em NIPS}, pages 3140--3148, 2014.

\bibitem{Li_transfer_2009}
Bin Li, Qiang Yang, and Xiangyang Xue.
\newblock Transfer learning for collaborative filtering via a rating-matrix
  generative model.
\newblock In {\em ICML}, pages 617--624, 2009.

\bibitem{Lichman:2013}
M.~Lichman.
\newblock {UCI} machine learning repository, 2013.

\bibitem{NIPS2010_4151}
Dahua Lin, Eric Grimson, and John~W. Fisher.
\newblock Construction of dependent dirichlet processes based on poisson
  processes.
\newblock In {\em NIPS}, pages 1396--1404. 2010.

\bibitem{linero2017bayesian}
Antonio~Ricardo Linero and Yun Yang.
\newblock Bayesian regression tree ensembles that adapt to smoothness and
  sparsity.
\newblock {\em arXiv preprint arXiv:1707.09461}, 2017.

\bibitem{nakano2014rectangular}
Masahiro Nakano, Katsuhiko Ishiguro, Akisato Kimura, Takeshi Yamada, and
  Naonori Ueda.
\newblock Rectangular tiling process.
\newblock In {\em ICML}, pages 361--369, 2014.

\bibitem{nowicki2001estimation}
Krzysztof Nowicki and Tom~A.B. Snijders.
\newblock Estimation and prediction for stochastic block structures.
\newblock {\em Journal of the American Statistical Association},
  96(455):1077--1087, 2001.

\bibitem{scikit-learn}
F.~Pedregosa, G.~Varoquaux, A.~Gramfort, V.~Michel, B.~Thirion, O.~Grisel,
  M.~Blondel, P.~Prettenhofer, R.~Weiss, V.~Dubourg, J.~Vanderplas, A.~Passos,
  D.~Cournapeau, M.~Brucher, M.~Perrot, and E.~Duchesnay.
\newblock Scikit-learn: Machine learning in {P}ython.
\newblock {\em Journal of Machine Learning Research}, 12:2825--2830, 2011.

\bibitem{porteous2008multi}
Ian Porteous, Evgeniy Bart, and Max Welling.
\newblock {Multi-HDP}: A non parametric {Bayesian} model for tensor
  factorization.
\newblock In {\em AAAI}, pages 1487--1490, 2008.

\bibitem{BART_r_package}
{Robert McCulloch, Rodney Sparapani, Robert Gramacy, Charles Spanbauer, Matthew
  Pratola}.
\newblock {\em BART}.
\newblock R Foundation for Statistical Computing, 2018.

\bibitem{roy2011thesis}
Daniel~M. Roy.
\newblock {\em Computability, Inference and Modeling in Probabilistic
  Programming}.
\newblock PhD thesis, MIT, 2011.

\bibitem{roy2007learning}
Daniel~M. Roy, Charles Kemp, Vikash Mansinghka, and Joshua~B. Tenenbaum.
\newblock Learning annotated hierarchies from relational data.
\newblock In {\em NIPS}, pages 1185--1192, 2007.

\bibitem{roy2009mondrian}
Daniel~M. Roy and Yee~Whye Teh.
\newblock The {Mondrian} process.
\newblock In {\em NIPS}, pages 1377--1384, 2009.

\bibitem{sisson05}
Scott~A. Sisson.
\newblock Trans-dimensional {Markov} chains: {A} decade of progress and future
  perspectives.
\newblock {\em Journal of the American Statistical Association},
  100:1077--1089, 2005.

\end{thebibliography}
\bibliographystyle{plain}

\end{document}